\documentclass[letterpaper, 10 pt, conference]{ieeeconf}  

\IEEEoverridecommandlockouts                              

\usepackage{lipsum}
\usepackage{color}
\usepackage[colorinlistoftodos]{todonotes}

\usepackage{algorithm}
\usepackage{algpseudocode}
\usepackage{pifont}

\usepackage{amsmath}
\usepackage{amssymb}
\usepackage{amsfonts}
\usepackage{mathtools}
\usepackage{bm}
\usepackage{mathrsfs}
\usepackage{eucal} 

\usepackage{graphicx}
\usepackage{epstopdf}
\usepackage[font=footnotesize]{caption}
\usepackage{subcaption}
\usepackage{wrapfig}
\usepackage{tikz}
\usepackage{tabularx}
\usepackage{placeins}
\usepackage{dblfloatfix}

\usepackage[hang,flushmargin]{footmisc}   

\usepackage{float}
\usepackage{placeins}
\usepackage{booktabs}
\usepackage{tabularx}

\newcommand*\rot{\rotatebox{66}}

\title{\LARGE \bf
Navigating Occluded Intersections with Autonomous Vehicles 
\\ using Deep Reinforcement Learning
}

\author{David Isele$^{1,3}$, Reza Rahimi$^{2}$, Akansel Cosgun$^{3}$, Kaushik Subramanian$^{4}$ and Kikuo Fujimura$^{3}$
\thanks{$^{1}$The University of Pennsylvania, $^{2}$The University of Virginia,}
\thanks{$^{3}$Honda Research Institute,$^{4}$The Georgia Institute of Technology}
}

\begin{document}

\maketitle
\thispagestyle{empty}
\pagestyle{empty}

\begin{abstract}

Providing an efficient strategy to navigate safely through unsignaled intersections is a difficult task that requires determining the intent of other drivers. We explore the effectiveness of Deep Reinforcement Learning to handle intersection problems. Using recent advances in Deep RL, we are able to learn policies that surpass the performance of a commonly-used heuristic approach in several metrics including task completion time and goal success rate and have limited ability to generalize. We then explore a system's ability to learn active sensing behaviors to enable navigating safely in the case of occlusions. Our analysis, provides insight into the intersection handling problem, the solutions learned by the network point out several shortcomings of current rule-based methods, and the failures of our current deep reinforcement learning system point to future research directions.



\end{abstract}

\section{INTRODUCTION}

One of the most challenging problems for autonomous vehicles is to handle unsignaled intersections in urban environments. To successfully navigate through an intersection, it is necessary to understand vehicle dynamics, interpret the intent of other drivers, resolve the blind regions in case of occlusion, and behave predictably so that other drivers can respond appropriately. Learning this behavior requires optimizing multiple conflicting objectives including safety, efficiency, and minimizing the disruption of traffic. The ability to perform optimally at traffic junctions can both extend the abilities of autonomous agents and increase safety through driver assistance when a human driver is in control.

Several strategies have already been applied to intersection handling, including cooperative \cite{hafner2013cooperative} and heuristic \cite{alonso2011autonomous} approaches. Cooperative approaches require vehicle-to-vehicle communication and thus are not scalable to general intersection handling. The current state of the art is a rule-based method based on time-to-collision (TTC) \cite{minderhoud2001extended, van1993time}, which is a widely used heuristic as a safety indicator in the automotive industry \cite{vogel2003comparison}. Variants of the TTC approach have been used for autonomous driving \cite{ferguson2008reasoning} and the DARPA urban challenge, where hand engineered hierarchical state machines were a popular approach to handle intersections \cite{darpa2, urmson2008autonomous}. TTC is currently the method we employ on our autonomous vehicle \cite{cosgun2017towards}.

While TTC has many benefits - it is relatively reliable, generates behavior that is easy to interpret, and can be tuned to reach a high level of safety - it also has limitations. First, the TTC models assume constant velocity, which ignores nearly all information concerning driver intent. This is problematic: in the DARPA Urban Challenge, one reason behind a collision between two autonomous cars was ``failure to anticipate vehicle intent'' \cite{darpa1}. Even with higher order dynamics, the often-unpredictable behavior of human drivers complicates the use of rule-based algorithms. Second, in many cases an agent using TTC is overly cautious, creating unnecessary delays. Third, TTC methods assume full knowledge of their surroundings limiting their ability to handle occlusions. While special behaviors can be coded to address specific cases, this requires an engineer to identify and manage a large possible number of scenarios. These reasons motivate our investigation of machine-learning based approaches for intersection handling in autonomous vehicles.

\begin{figure}[t]
\centering
  \includegraphics[width=0.5\textwidth]{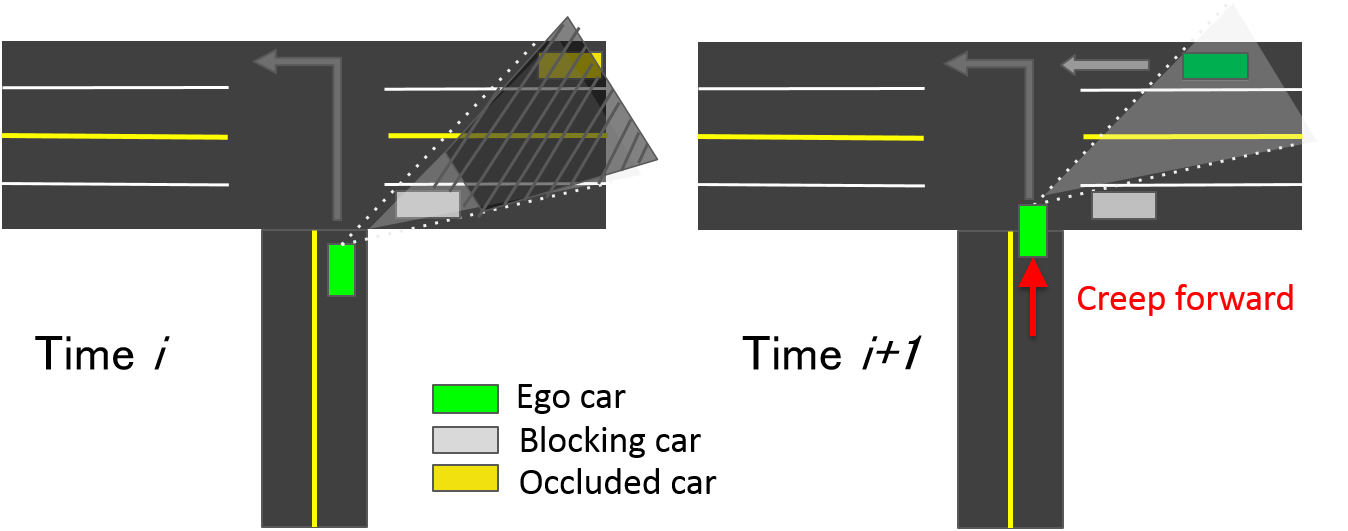}
  \caption{Using creeping behavior to actively sense occluded obstacles. The objective is to determine the acceleration profile along the path while safely avoiding collisions.}
  \label{fig:sumo}
  \vspace{-0.2cm}
\end{figure}

Several machine learning based approaches have been used for the intersection handling, such as imitation learning, online planning, and offline learning. In imitation learning, the policy is learned from a human driver \cite{bojarski2016end}, however  this policy does not offer a solution if the agent finds itself in a state that is not part of the training data. Online planners compute the best action to take by simulating the future states from the current time step. Online planners based on partially observable Monte Carlo Planning (POMCP) have been shown to handle intersections \cite{bouton2017belief}, but rely on the existence of an accurate generative model. Offline learning tackles the intersection problem, often by using Markov Decision Processes (MDP) in the back-end \cite{song2016intention,brechtel2014probabilistic}. 

In this paper, we explore the use of Deep Q-Networks (DQNs) \cite{mnih2013playing,wang2015dueling} 
for intersection handling with a focus on acting in the presence of partial knowledge. We view exploratory actions as a complementary approach to existing partially observable Deep RL methods which use histories to approach the problem of partial observability \cite{zhang2016learning,heess2015memory}. 
Figure \ref{fig:sumo} shows an autonomous vehicle at an unsignaled intersection where an occlusion blocks the view of the road. In order to navigate safely, the agent must first take an exploratory action to better understand the environment. 

The first contribution of this paper is demonstrating the effectiveness of deep learning techniques on the intersection handling problem. We explore several network designs and present two we found to work well. The top performing system works by identifying the time to start accelerating. 
As neither TTC nor DQNs perform optimally in all cases, our analysis has interest beyond the specific choice of learner as we identify scenarios where TTC can be improved and suggests methods that could improve it.

The second contribution is the analysis of exploratory actions (creeping behaviors in specific) as a means of improving safety on autonomous vehicles. 
We show that occlusions create a need for exploratory actions and we show that deep reinforcement learning agents are able to discover these behaviors. 


\begin{figure*}[thpb!]
    \centering
    \vspace{8pt}
    \hspace{-20pt}
    \begin{subfigure}[b]{1.1in}
    	\includegraphics[height=.75\textwidth]{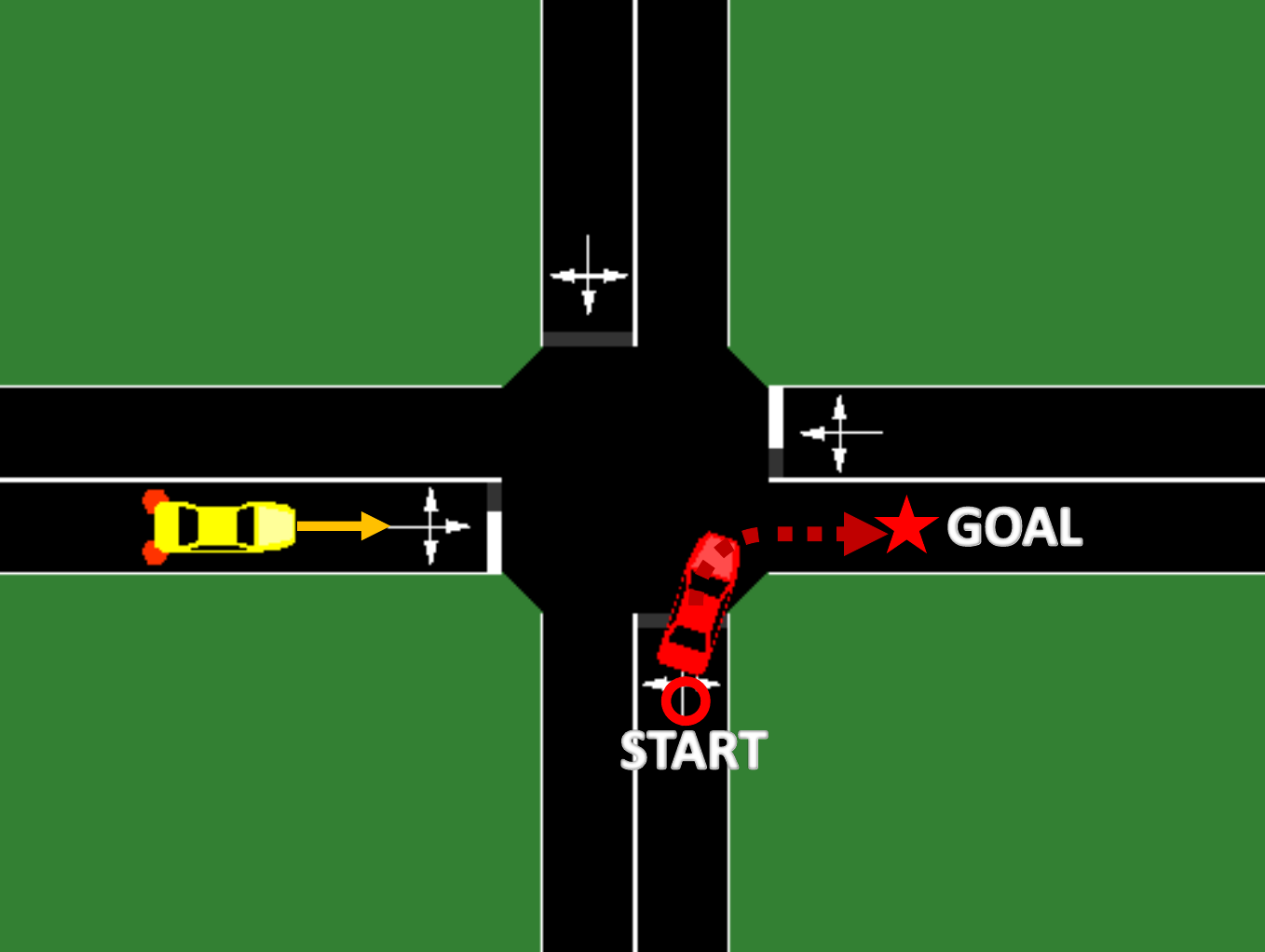} 
        \caption{\emph{Right}}
        \label{fig:scenarios_right}
    \end{subfigure}
    \begin{subfigure}[b]{1.1in}
    	\includegraphics[height=.75\textwidth]{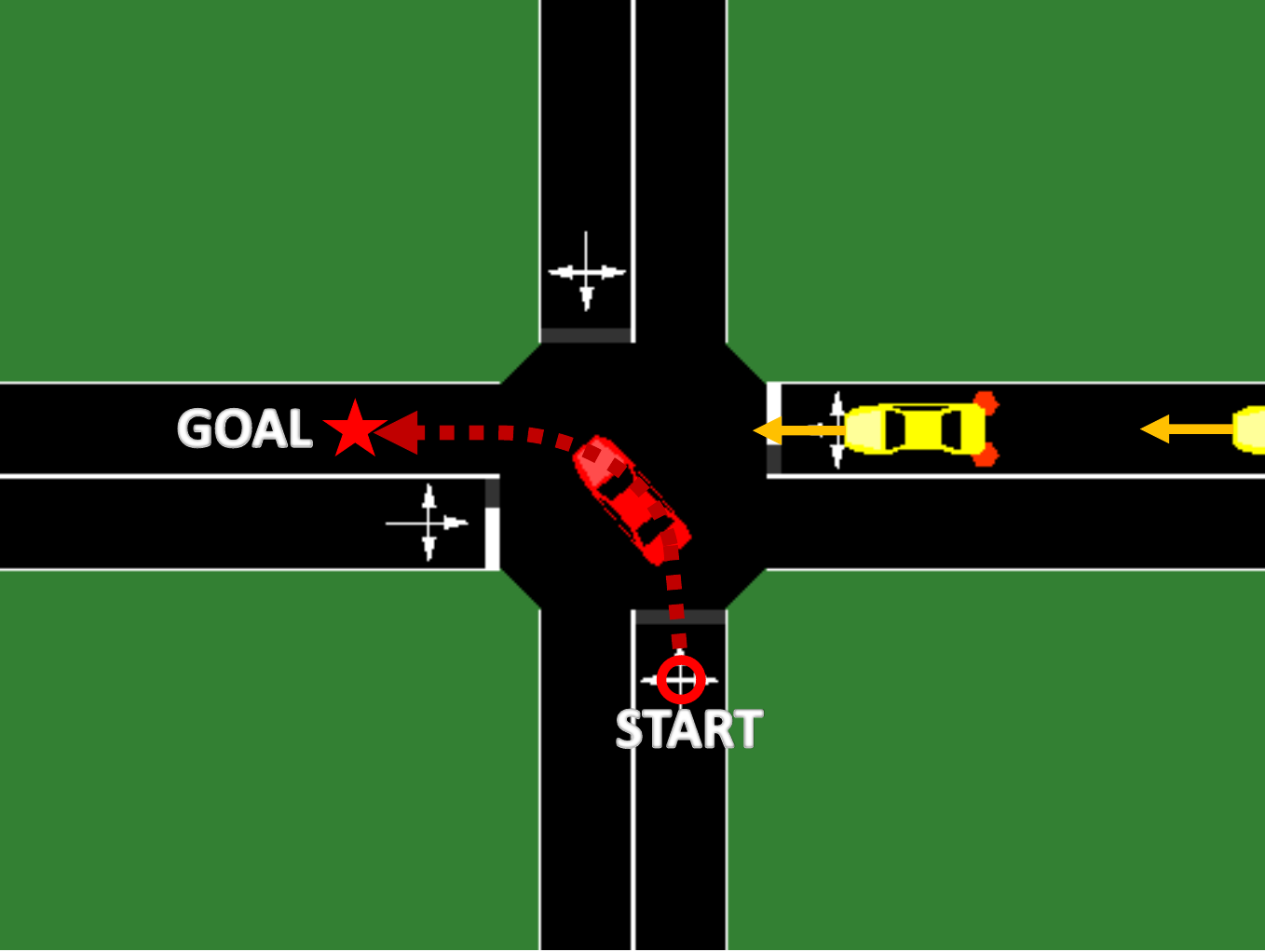}
        \caption{\emph{Left}}
        \label{fig:scenarios_left}
    \end{subfigure}
    \begin{subfigure}[b]{1.1in}
    	\includegraphics[height=.75\textwidth]{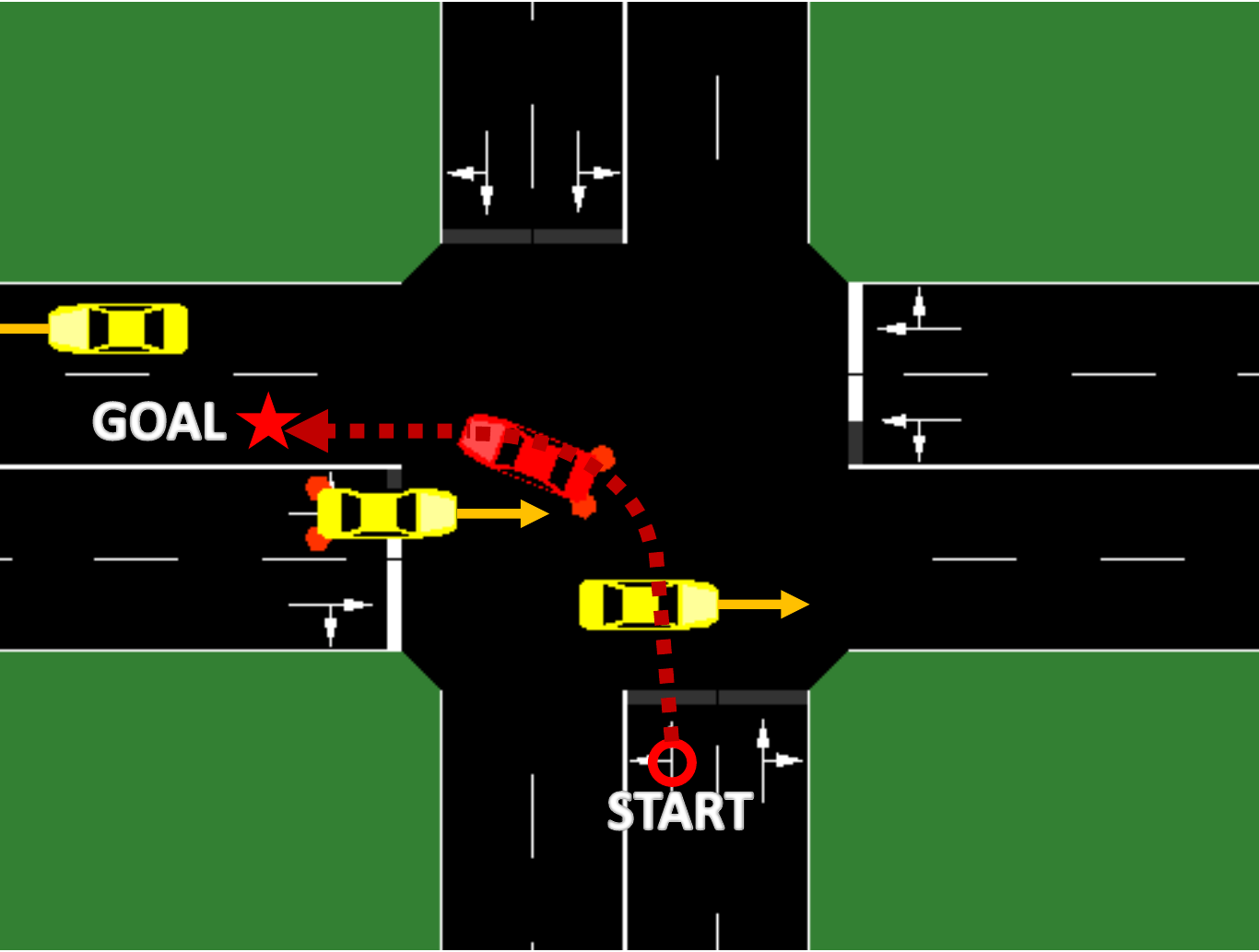}
        \caption{\emph{Left2}}
        \label{fig:scenarios_left2}
    \end{subfigure}
    \hspace{-4pt}
    \begin{subfigure}[b]{1.1in} 
       \includegraphics[height=.75\textwidth]{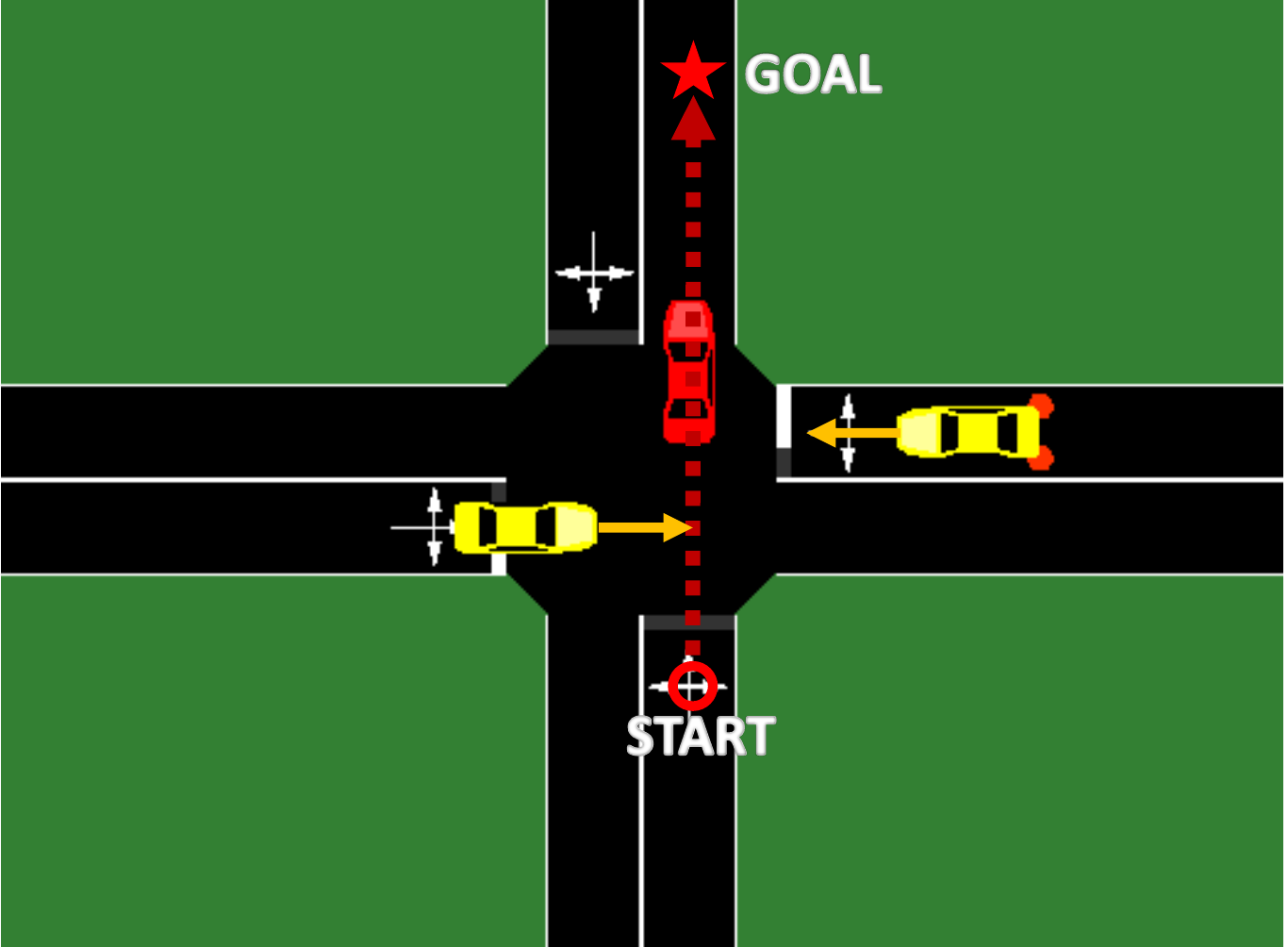}
        \caption{\emph{Forward}}
        \label{fig:scenarios_forward}
    \end{subfigure}
    \hspace{1pt}
    \begin{subfigure}[b]{1.1in}
    	\includegraphics[height=.75\textwidth]{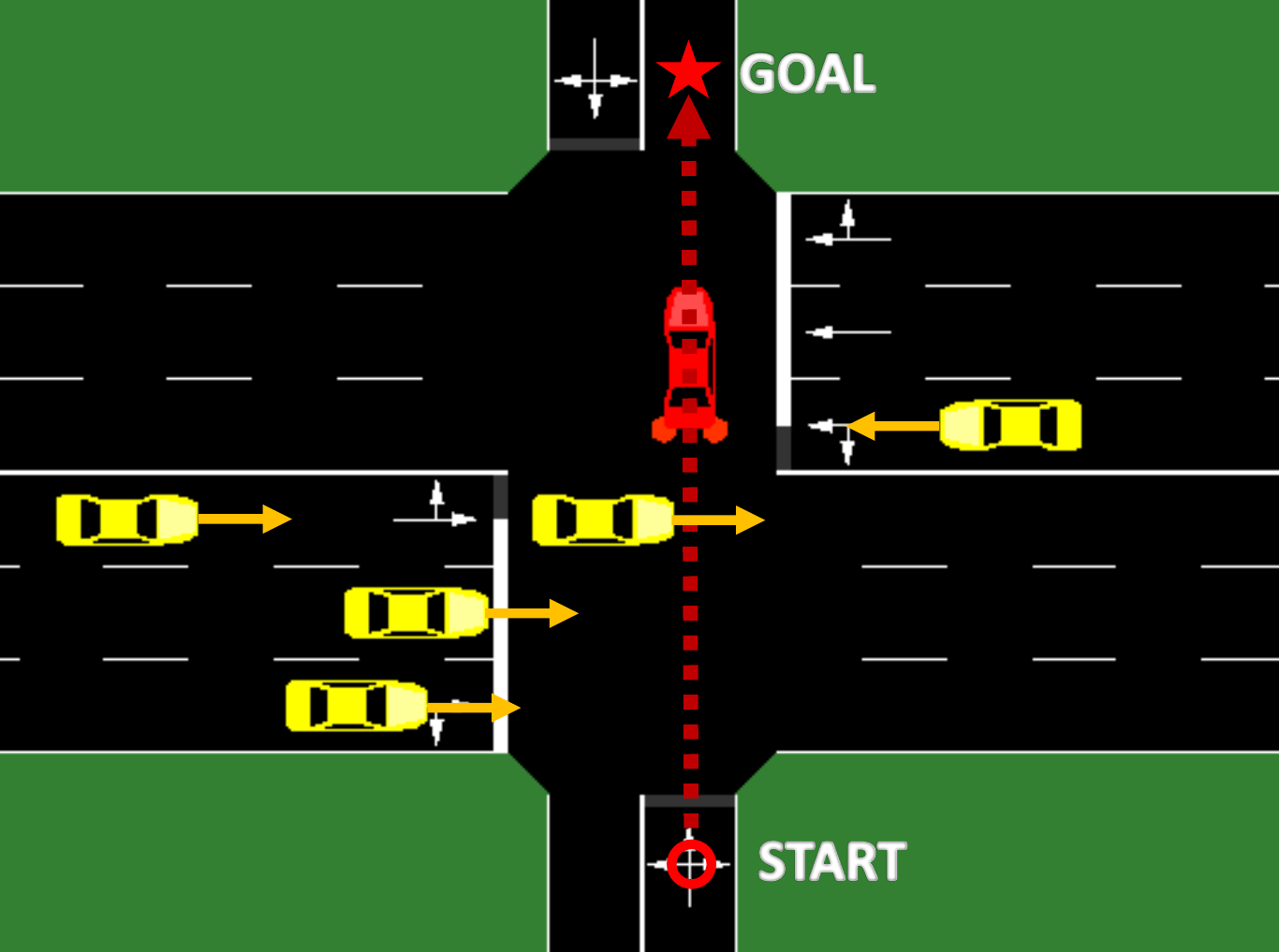}
        \caption{\emph{Challenge}}
        \label{fig:scenarios_challenge}
    \end{subfigure}
        \caption{Visualizations of intersection tasks used for our experiments.}\label{fig:scenarios} 
        \vspace{-10pt}
\end{figure*}

\section{APPROACH}
\label{sec:approach}

We view intersection handling as a reinforcement learning problem, and use a Deep Q-Network (DQN) to learn the state-action value Q-function. 

\subsection{Reinforcement Learning}
In the reinforcement learning framework, at time $t$ an agent in state $s_t$ takes an action $a_t$ according to the policy $\pi$. The agent transitions to the state $s_{t+1}$, and receives a reward $r_t$. The sequence of states, actions, and rewards is given as a trajectory \mbox{$\mathbf{\tau} = \{(\mathbf{s}_1, a_1, r_1), \ldots, (\mathbf{s}_T, a_T, r_T)\}$} over a horizon $T$. 

A reinforcement learning task is typically formulated as a Markov Decision Process (MDP) $\langle \mathcal{S}, \mathcal{A}, P, \mathcal{R}, \gamma  \rangle$, where $\mathcal{S}$ is the set of states, and $\mathcal{A}$ is the set of actions that the agent may execute. MDPs follow the Markov assumption that the probability of transitioning to a new state, given the current state and action, is independent of all previous states and actions $p(\mathbf{s}_{t+1}|\mathbf{s}_{t}, a_{t}, \dots, \mathbf{s}_{0}, a_{0}) = p(\mathbf{s}_{t+1}|\mathbf{s}_{t}, a_{t})$. The state transition probability \mbox{$P: \mathcal{S} \times \mathcal{A} \times \mathcal{S} \rightarrow [0,1]$} describes the systems dynamics, the reward function $\mathcal{R}: \mathcal{S} \times \mathcal{A} \times \mathcal{S} \rightarrow \mathbb{R} $ gives the real valued reward for a given time step, and $\gamma \in (0,1]$ is a discount factor that adds preference for earlier rewards and provides stability in the case of infinite time horizons.

The goal of reinforcement learning is to choose the sequence of actions starting at time $t$ that maximize the expected return,  $R_t = \sum_{k=0}^T \gamma^t r_{t+k}$. We use Q-learning to perform this optimization.



\subsection{Q-learning}
In Q-learning \cite{watkins1992q}, the action value function $Q^\pi(s,a)$ is the expected return $\mathbb{E}[R_t|s_t=s,a_t=a]$ for a state-action pair following a policy $\pi$. Given an optimal value function $Q^*(s,a)$ the optimal policy can be inferred by selecting the action with maximum value $\max_a Q^*(s,a)$ at every time step. 

In Deep Q-learning \cite{mnih2013playing}, the optimal value function is approximated with a neural network $Q^*(s,a)\approx Q(s,a;\theta)$ with parameters $\theta$. The action value function is learned by iteratively minimizing the error between the expected return and the state-action value predicted by the network

\vspace{-0.4cm}

\begin{eqnarray}
\mathcal{L(\theta)} = \bigg( \mathbb{E}[R_t|s_t=s,a_t=a] - Q(s,a;\theta)\bigg)^2 \enspace .
\end{eqnarray}
Since in practice the true return is not known, it is often approximated by the one-step return 
\begin{eqnarray}
\mathbb{E}[R_t|s_t=s,a_t=a] \approx r_t + \gamma \max_{a_{t+1}}Q(s_{t+1},a_{t+1};\theta)  \enspace .
\end{eqnarray}

\subsection{Action Representations}
Autonomous vehicles use a suite of sensors, and allow for planning at multiple levels of abstraction. This allows for a wide variety of state and action representations. An exploratory search of representations showed that the selection of the representation had a significant impact on the agent's ability to learn. In this paper, we focus on three representations we found to work well.

\paragraph{Time-to-Go}
In the Time-to-Go representation the desired path is provided to the agent and the agent determines the timing of departure through a sequence of actions to wait or go. Every \emph{wait} action is followed by another wait or go decision, meaning every trajectory is a series of wait decisions terminating in a go decision, and the agent is not allowed to wait after the go action has been selected. 

\paragraph{Sequential Actions}
In the sequential action representation, the desired path is provided to the agent and the agent determines to accelerate, decelerate, or maintain constant velocity at every point in time along the desired path. 

The sequential scenario allows for more complex behaviors: the agent could potential slow down half way through the intersection and wait for oncoming traffic to pass. The Time-to-Go representation more closely compares to TTC, placing importance on the time of departure.
By analyzing the sequential action representation, we can observe if there is a benefit to allowing more complex behaviors. The Time-to-Go representation focuses on the departure time, allowing us to specifically probe how changes in departure time can affect performance.

\paragraph{Creep-and-Go}


A third action set is constructed as a hybrid between sequential actions and Time-to-Go. Creep-and-Go involves three actions: wait, move forward slowly, and go. Like Time-to-Go, once a go action is selected the agent continues all the way through the intersection. This action configuration allows the agent to choose between moving up slowly and stopping before finally choosing a go action. 

Creep-and-Go offers easier interpretability and ease of learning in a restricted action space like Time-to-Go, but makes exploratory actions available to the agent like in the sequential action system.

\subsection{State Representation}

A bird's eye view of space is discretized into a grid in Cartesian coordinates. Every car in the space is represented by its heading angle, velocity, and an indicator term that is $1$ when a car occupies the position in the grid and $0$ otherwise. Preliminary studies showed that real valued representations perform better, which we hypothesize is due to the reduced state space being easier to learn and the real values making it easier for the system to generalize.   


In the occluded setting the representation is increased to include an indicator to identify if the cell is occluded, and the offset of the car's $x$ and $y$ offset to the pixel boundary to help assuage discretization affects. 


\FloatBarrier

\section{EXPERIMENTS}
\label{sec:experiments}
We first train two different DQNs (Sequential Actions and Time-to-Go) on a variety of intersection scenarios and compare the performance against the heuristic Time-to-Collision (TTC) algorithm \cite{minderhoud2001extended, van1993time, cosgun2017towards}. We then investigate a DQNs ability to learn exploratory behaviors in order to navigate occluded intersections.

\subsection{Algorithm Comparison Study}

Experiments were run using the Sumo simulator \cite{sumo}, which is an open source traffic simulation package. 
While we are presently restricting our reinforcement learning experiments to simulation for safety reasons, we verify that the representations generated from the simulator are qualitative similar to representations created from data collected from a real vehicle and have conducted initial investigations into transferring policies from simulation to a real vehicle \cite{isele2017transferring}. We collected data from an autonomous vehicle in Mountain View, California, at an unsigned T-junction, similar to the \emph{Left} scenario. A point cloud, obtained by a combination of six IBEO Lidar sensors, is first pre-processed to remove points that reside outside the road boundaries. A clustering method with hand-tuned geometric thresholds is used for vehicle detection. Each vehicle is tracked by a separate particle filter. 

To simulate traffic in Sumo, users have control over the types of vehicles, road paths, vehicle density, and departure times. Traffic cars follow the Intelligent Driver Model (IDM) \cite{idm-def} to control their motion. In Sumo, randomness is simulated by varying the speed distribution of the vehicles and by using parameters that control driver imperfection (based on the Krauss stochastic driving model \cite{krauss1998sumo}). The simulator runs based on a predefined time interval which controls the length of every step.
Our first set of experiments were to compare action representations against the TTC baseline.
We ran experiments using five different intersection scenarios: \emph{Right}, \emph{Left}, \emph{Left2}, \emph{Forward} and a \emph{Challenge}. Each of these scenarios is depicted in Figure \ref{fig:scenarios}.
The \emph{Right} scenario involves making a right turn, the \emph{Forward} scenario involves crossing the intersection, the \emph{Left} scenario involves making a left turn, the \emph{Left2} scenario involves making a left turn across two lanes, and the \emph{Challenge} scenario involves crossing a six-lane intersection with increased traffic density. The challenge scenario was intended to offer the sequential action model opportunity to benefit from more dynamic actions.

Each lane has a 45 mile per hour (20 m/s) maximum speed. The car begins from a stopped position. Each time step is equal to 0.2 seconds. The maximum number of steps per trial is capped at 100 steps which is equivalent to 20 seconds. The traffic density is set by the probability that a vehicle will be emitted randomly per second. We use 0.2 for all scenarios except the challenge scenario where it is set to 0.7.

We evaluate each method according to four metrics and run 10,000 trials of each scenario to collect our statistics. We use the following metrics:
\begin{itemize}
\item \textbf{Percentage of successes:} the percentage of the runs where the car successfully reached the goal. This metric takes into account both collisions and time-outs.
\item \textbf{Percentage of collisions:} a measure of the safety of the method. 
\item \textbf{Average time:} how long it takes a successful trial to run to completion.
\item \textbf{Average braking time:} the amount of time other cars in the simulator are braking. This can be seen as a measure of how disruptive the autonomous car is to traffic.
\end{itemize}


The networks architectures were optimized for sequential DQN and Time-to-Go DQN separately. 
For the sequential action DQN, space is represented as a $5 \times 11$ grid discretizing 0 to 20 meters in front of the car and $\pm 90$ meters to the left and right of the car. Each spatial pixel, if occupied, contains the normalized real valued heading angles, velocity, and calculated time to collision. Exploratory studies found that this spatial representation outperformed higher dimensional representations with finer granularity. It was also found that real representations of the heading angle, velocity, and time to collision outperformed discretized versions. We hypothesize that this is because the real values allow for greater generalization. 

For the Time-to-Go DQN, space is represented as an $18 \times 26$ grid. Unlike the sequential action DQN, this representation does not use the calculated time to collision for each car.  

The sequential action network has 12 outputs correspond to three actions (accelerate, decelerate, maintain velocity) at four time scales (1, 2, 4, and 8 time steps) following the dynamic frame-skipping strategies of Lakshminarayanan et al. \cite{lakshminarayanan2016dynamic}. 
The final output layer for Time-to-Go uses five outputs: a single \emph{go} action, and a \emph{wait} action at four time scales (1, 2, 4, and 8 time steps). Both networks are optimized using the RMSProp algorithm \cite{tieleman2012lecture}. 


The sequential action network was slower to learn and performed considerably below the Time-to-Go.
To put the performance of the sequential network on par with Time-to-Go an additional feature denoting the time to collision for each car was added to the state representation and the network was allowed to train longer.
Each sequential action scenario was trained on one million simulations. Each Time-to-go scenario was trained on 250 thousand simulations.

For the reward, we used $+1$ for successfully navigating the intersection, $-10$ for a collision, and $-0.01$ step cost. 

We additionally run experiments to test generalization. This is done by testing a policy trained in one scenario on other scenarios. Each policy is directly copied, and we do not do any additional training on the new scenarios. For generalization we used fewer training simulations (25,000) to prevent over-fitting.

\subsection{Experiments with Occlusion}
\vspace{-10pt}
\begin{figure}[thpb!]
    \centering
    	\includegraphics[clip,trim={20pt 0pt 20pt 0pt},height=.065\textwidth]{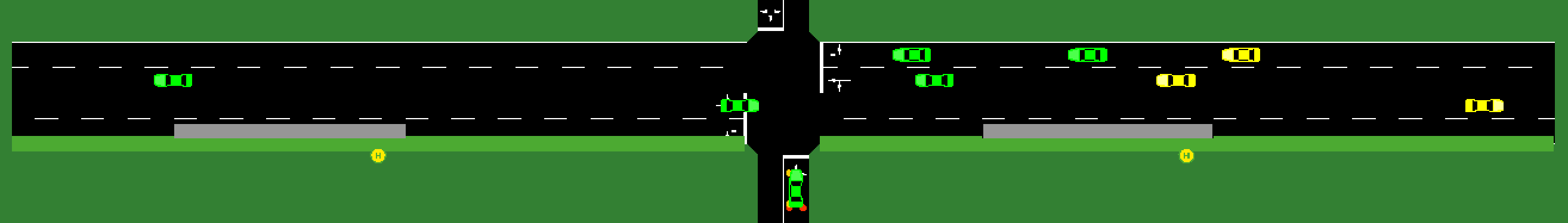}

\vspace{8pt}
       \includegraphics[clip,trim={50pt 0pt 20pt 0pt},height=.065\textwidth]{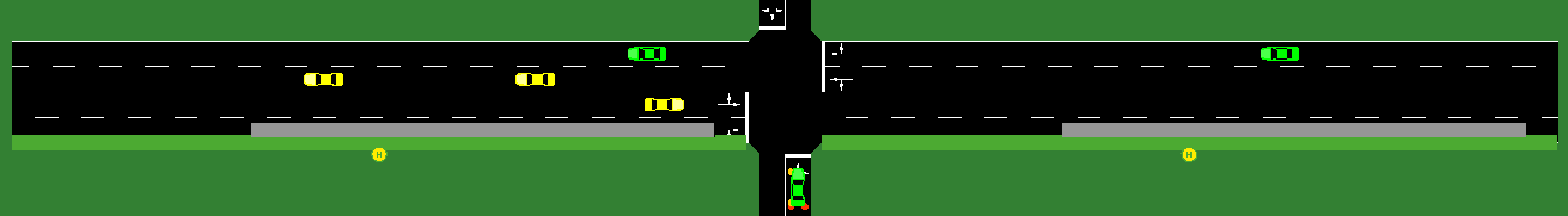}

\vspace{8pt}
    	\includegraphics[clip,trim={51pt 0pt 20pt 0pt},height=.065\textwidth]{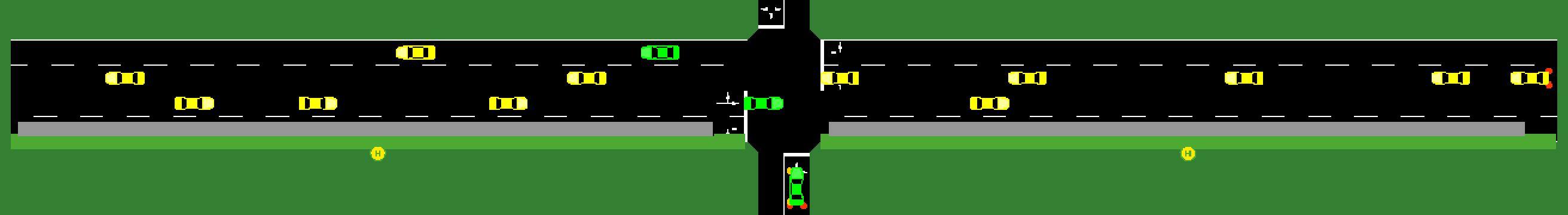}
        \caption{Visualizations of occlusion experiments. Occluding objects are gray. Occluded cars are shown in yellow, and visible cars are shown in green.}\label{fig:scenarios} 
\end{figure}
Our second set of experiments investigates handling intersections with occlusions. We consider an intersection setup where the goal is to make a left turn across two lanes in the presence of occlusions. The closest lane contains objects that obstruct the vehicles view. 
The visible area is modeled using a ray tracing approach to find the occluded cells in the grid. 
We do not add the information of the cars that are in an occluded cell.

Occlusions are randomly selected to appear on the left, right or both sides of the intersection. The occlusions vary in dimensions with lengths$=\{30,60,90\}$ meters and widths$=\{0.5,1,2\}$ meters. Each occlusion is positioned a distance of $\{51,61,91\}$ meters from the left most boundary of that side. For our occlusion experiments we switch to a dueling network architecture \cite{wang2015dueling} with prioritized replay \cite{schaul2015prioritized} which we found less sensitive to tuning and we doubled the traffic density to make differences in approaches more apparent. 
To allow more time to creep forward and find an opening in the denser traffic we increased the time limit to 60 seconds or 300 simulation steps.

To emphasize the importance of localizing cars in the occlusion setting, non-ego cars are not allowed to decelerate. This removes much of the network's ability to infer driver intent that was present in the first experiment, but makes missing the presence of a car more likely to result in a collision. Because traffic cars do not brake for the ego-car we do not measure braking time in the occlusion experiments.



\begin{table}[htbp!]
\caption{Comparison of Different Algorithms}
\begin{center}
\begin{tabular}{ccccccc}
\toprule 
    Scenario & Metric			 & \rot{\emph{TTC}} & \rot{\emph{\scriptsize DQN-Sequential}} & \rot{\emph{\scriptsize DQN-Time-to-Go}}  \\           
    \midrule
    \emph{Right} 
    & $\%$ Success			    & 99.61 & 99.5  & \textbf{99.96} \\
    & $\%$ Collisions 			& \textbf{0.0}   & 0.47  & 0.04  \\
    & Avg. Time 				& 6.46s & 5.47s & \textbf{4.63s} \\
    & Avg. Brake 			    & \textbf{0.31s} & 0.88s & 0.45s \\
    \midrule
    \emph{Left} 
    & $\%$ Success			    & 99.7  & \textbf{99.99} & \textbf{99.99}  \\
    & $\%$ Collisions           & \textbf{0.0}   & \textbf{0.0}  & 0.01   \\
    & Avg. Time                 & 6.97s & 5.26s & \textbf{5.24s}  \\
    & Avg. Brake                & 0.52s & \textbf{0.38s}  & 0.46s  \\
  \midrule
  \emph{Left2} 
  & $\%$ Success			    &  99.42  & 99.79 & \textbf{99.99} \\
  & $\%$ Collisions             &  \textbf{0.0}    & 0.11  & 0.01  \\
  & Avg. Time                   &  7.59s  & 7.13s & \textbf{5.40s} \\
  & Avg. Brake                  &  0.21s  & 0.22s & \textbf{0.20s} \\
  \midrule
  \emph{Forward}   
  & $\%$ Success			    &  \textbf{99.91}  & 99.76 & 99.78 \\
  & $\%$ Collisions             &  \textbf{0.0}    & 0.14  & 0.01  \\
  & Avg. Time                   &  6.19s  & \textbf{4.40s} & 4.63s \\
  & Avg. Brake                  &  0.57s  & 0.61s & \textbf{0.48s} \\
  \midrule
  \emph{Challenge}  
  & $\%$ Success			    & 39.2    & 82.97 & \textbf{98.46} \\
  & $\%$ Collisions             & \textbf{0.0}     & 1.37  & 0.84  \\
  & Avg. Time                   & 12.55s  & 9.94s & \textbf{7.94s} \\
  & Avg. Brake                  & \textbf{1.65s}   & 1.94s & 1.98s \\
\bottomrule
\end{tabular}
\end{center}
\label{table:comparison}
\end{table}

\vspace{-0.3cm}

\section{RESULTS}
\label{sec:results}

Table \ref{table:comparison} shows the results from our first set of experiments. To yield the best results for TTC, the threshold for each scenario is chosen as the lowest that achieve zero collisions. 
As a result, The TTC method did not have a collision in any of the scenarios. All other methods had non-zero collision rates for all scenarios, except DQN-Sequential for the \emph{Left} scenario. Among DQN methods, DQN Time-to-Go had substantially lower collision rate than DQN-sequential. For scenarios except \emph{Challenge}, DQN Time-to-Go had only 7 collisions in a total of 40,000 simulations. 


We see that both DQN methods are substantially more efficient at reaching the goal than TTC. DQN Time-to-Go has the best task completion time in all scenarios, except \emph{Forward}, where DQN-Sequential is faster. On average, DQN Time-to-Go was $28\%$ faster at reaching goal than TTC, whereas DQN Sequential was $19\%$ faster than TTC. Therefore, the DQN methods have potential to reduce traffic congestion due to their efficiency navigating intersections.

Braking time, a measure of how disruptive the car is to other traffic, is comparable for all methods. Because of this, we did not find conclusive evidence of DQN methods being more aggressive than TTC, despite DQN methods posting non-zero collision rates.

DQN Time-to-Go has the highest success rates in all experiments, except one, where its success rate is only marginally lower than TTC. Both DQN methods produce sound policies, evidenced by the ego car reaching the goal at least $99.5\%$ of the time for all scenarios except \emph{Challenge}. 

An interesting result was that TTC did not reach the goal the majority of the time in the \emph{Challenge} scenario, reaching the goal only $39.2\%$ of the time, and posting a success rate only slightly more than Random ($29.9\%$). For the same scenario, DQN Time-to-Go reached the goal $98.46\%$ of the time, significantly outperforming other methods. We offer more insight on these results in Section \ref{sec:qualitative}.

\begin{figure}[thpb!]
    \centering
    \vspace{-10pt}
    \hspace{-20pt}
    	\includegraphics[width=0.35\textwidth]{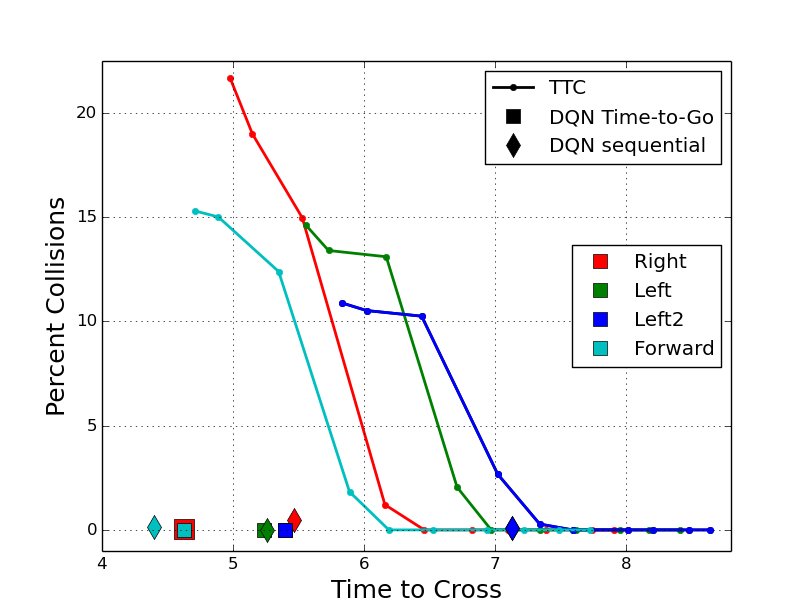}
        \caption{Trade-off between the time to cross and collision rate as the TTC threshold is varied. Note that performance of the DQN dominates in every case. The challenge scenario is excluded for scale reasons, but the results are similar.}
\label{fig:pareto}
\end{figure}


\begin{table}[hbpt!]
\caption{Comparison of Different Algorithms on Occlusion}
\begin{center}
\begin{tabular}{ccccccc}
\toprule 
    \rot{Occlusion Length} & \rot{Metric} & \rot{\emph{TTC w creep}} & \rot{\emph{DQN-Time-to-Go}} & \rot{\emph{\scriptsize DQN-Creep}} & \rot{\emph{\scriptsize DQN-Sequential}}  \\   
    \midrule
    \emph{30m} 
    & $\%$ Success			    & 84.1 & 70.7 & \textbf{92.8}  & 92.6 \\
    & $\%$ Collisions 			& \textbf{0.0} & 29.2  & 7.1  & 7.3  \\
    & Avg. Time 				& 28.6s  & 12.0s & 11.7s & \textbf{9.6s} \\
    \midrule
    \emph{60m} 
    & $\%$ Success			   & 83.1 & 57.2  & \textbf{92.9} & 88.9  \\
    & $\%$ Collisions          & \textbf{0.0} & 36.65 & 7.1  & 11.05   \\
    & Avg. Time                & 28.9s & 21.3s & 16.0 & \textbf{9.91s}  \\
  \midrule
  \emph{90m} 
  & $\%$ Success			   & 87.6 & 26.3  & \textbf{96.3} & 86.2 \\
  & $\%$ Collisions            &  \textbf{0.0}  & 27.9 & 3.7  & 13.8  \\
  & Avg. Time                  & 28.7s &  30.8s  & 16.0s & \textbf{10.25s} \\
\bottomrule
\end{tabular}
\vspace{-10pt}
\end{center}
\label{table:occlusion}
\end{table}


While the DQNs are substantially more efficient, they are seldom able to minimize the number of collisions as successfully as TTC. This is due to the fact that TTC has a tunable parameter that adjusts the safety margin and we tune TTC to the lowest threshold that gives zero collisions.  

Comparing the DQN's performance against the TTC curve as we trade off speed vs. safety (Figure \ref{fig:pareto}), we see that in every instance the DQN's performance dominates the performance of TTC. This suggests that it is possible to design an algorithm that has zero collision rate, but with better performance metrics than TTC.

\subsection{Safety, Generalization, and Transfer}

While the DQNs outperform TTC in terms of success rates, being able to achieve a zero percent collision rate is very important. The difficulty in shaping the reward to achieve zero collisions is a known problem with gradient based learning approaches \cite{shalev2016safe}. 
Seamlessly integrating safety constraints into the learning process is the focus on ongoing research \cite{gillula2013reducing,alshiekh2017safe}. However, unconstrained exploration is a useful tool for development - our experiments produced a policy where the autonomous vehicle aims to be very close behind a passing car in order to be as efficient as possible. This differed from the authors' natural driving, and appears to be a useful strategy for dense traffic. However, it is important to moderate the aggressiveness - applying the strategy too aggressively can create difficulty in interpreting intention and disrupt other drivers.      

While we are examining safety constraints in our ongoing research, in this work we conducted an initial investigation into safety concerning how the system generalizes to out-of-sample data. 
To do this, we run the network trained in one scenario on every other scenario, directly copying the policy. The results are shown in Table \ref{table:transfer_gonogo}.
 
\begin{table}[htp]
\caption{Transfer Performance for DQN Time-to-Go}
\begin{center}
\begin{tabular}{cccccccc}
\toprule 
    Scenario &  & \multicolumn{5}{c}{Training Method \vspace{-0.5cm}}\\    
    \cr & & \rot{\emph{Right}} & \rot{\emph{Left}} & \rot{\emph{Left2}} & \rot{\emph{Forward}}  & \rot{\emph{Challenge}}   \\        
    \midrule
    \emph{Right} 
    &	& \textbf{99.0} & 98.5   & 81.1  & 98.1 & 74.1 \\

    \emph{Left} 
    &    & 87.7 & \textbf{96.3} & 76.8   & 95.1 & 66.2 \\
  
    \emph{Left2} 
    & 	 & 70.1 & 76.3  & \textbf{91.7} & 74.5 & 74.6 \\
    
    \emph{Forward}    
    & 	 & 87.2 & 97.2  & 79.0  &\textbf{97.6} & 69.9 \\
    
    \emph{Challenge}    
    & 	 & 58.7 & 60.7  & 72.3  & 62.4  & \textbf{78.5} \\
\bottomrule
\end{tabular}
\end{center}
\label{table:transfer_gonogo}
\vspace{-10pt}
\end{table}

For both the sequential and Time-to-Go DQNs we see that the networks trained on the \emph{Left} scenario transfer better to the other single lane scenarios: \emph{Right} and \emph{Forward}. Similarly, the networks trained on the \emph{Left2} scenario transferred well to the other multi-lane setting: \emph{Challenge}. We believe this is largely due to the way the state is being represented. 
Notably, no method transfers well to all tasks. For a deeper investigation of how training on multiple tasks can increase a systems ability to generalize, we refers the readers to \cite{isele2018aaai}.


\subsection{Occlusion}

Table \ref{table:occlusion} shows the results of our occlusion experiments. 
A network using sequential actions can reach the point of full visibility fastest as therefore tends to be faster than the creeping approach, however the restricted action space does make creeping easier to learn and leads to a higher success rate. The creep forward behavior also brings the car closer to the goal when the intersection is blocked, as a result the methods without the creeping behavior (TTC and Time-to-Go) tend to take longer on average. 

We see that TTC without creeping incorrectly assumes the road is clear and results in collisions. Even though TTC with creeping behavior has no collisions, it has a high percentage of timeouts. Notably the DQN is able to learn the specialized behavior and perform it more efficiently than TTC. 

\subsection{Qualitative Analysis}
\label{sec:qualitative}

Comparing trials of the learned DQN networks and TTC, the DQN strategies take into account predictive behavior of the traffic. The DQNs can accurately predict that traffic in distant lanes will have passed by the time the ego car arrives at the lane. Also the DQN driver can anticipate whether oncoming traffic will have sufficient time to brake or not. The few collisions seem to relate to discretization effects, where the car nearly misses the oncoming traffic.

In contrast, TTC does not leave until all cars have cleared its path. In addition, TTC leaves a sufficient safety margin for oncoming cars in distant lanes, since the same safety margin is used, the gap gets exaggerated in closer lanes. As a result, TTC often waits until the road is completely clear, missing many opportunities to cross. 

We see that selecting the departure time offers sufficient opportunity to improve over TTC without the need to incorporate the added complexity of allowing for dynamic acceleration and deceleration behavior. This holds even for the \emph{Challenge} scenario. The Time-to-Go DQN often chooses departures when other cars are either in or approaching the intersection, correctly predicting that they will be clear by the time the car reaches that position. However, using only departure time becomes ineffective once the systems encounters occlusions. Without the ability to move to a more favorable vantage point, both TTC and Time-to-Go regularly encounter collisions. This study identifies occlusions as a case where exploratory behaviors are needed and shows that DQNs are capable of learning them. 


\section{CONCLUSIONS}
\label{sec:conclusion}


We showed a first system that uses Deep Q-Networks for the specific problem of intersection handling and show that it is capable of learning exploratory behaviors to more fully understand the scene. In addition to having better task efficiency and success rates than the rule based method, we identify occlusions as a fail case for rule based methods that lack hand-coded accommodations. 
While networks can learn both departure times and creeping behaviors, and have some ability to generalize to novel domains and out-of-sample data, they do occasionally result in collisions. Consequently more research is required to increase robustness.




\bibliographystyle{IEEEtran}
\small{
\bibliography{refs}
}
\end{document}